%% file: AI_ComMag.tex
\documentclass[10pt, conference, letterpaper]{IEEEtran}

\usepackage{color,overpic,cite}
\usepackage[lofdepth,lotdepth]{subfig}
\usepackage{graphicx}
\usepackage{epstopdf}
\usepackage{pdfsync}
\usepackage{url}
\usepackage{amsmath,amsfonts,amssymb,amsthm,bbm,bm,comment,multirow}
\usepackage{algorithm}
\usepackage{algorithmic}

\newcommand{\eat}[1]{{}}

\author{Spyridon Vassilaras, Luigi Vigneri, Nikolaos Liakopoulos, Georgios S. Paschos, \\ Apostolos Destounis, Thrasyvoulos Spyropoulos, and M\'{e}rouane Debbah}

\title{Problem-Adapted Artificial Intelligence for Online Network Optimization}

\begin{document}
\maketitle

\begin{abstract}
Future 5G wireless networks will rely on agile and automated network management, where the usage of diverse resources must be jointly optimized with surgical accuracy. A number of key wireless network functionalities (e.g., traffic steering, power control) give rise to hard optimization problems. What is more, high spatio-temporal traffic variability coupled with the need to satisfy strict per slice/service SLAs in modern networks, suggest that these  problems must be constantly (re-)solved, to maintain close-to-optimal performance. To this end, we propose the framework of \emph{Online Network Optimization} (ONO), which seeks to maintain both agile \emph{and} efficient control over time, using an arsenal of data-driven, online learning, and AI-based techniques. Since the mathematical tools and the studied regimes vary widely among these methodologies, a theoretical comparison is often out of reach. Therefore, the important question ``\emph{what is the right ONO technique?}'' remains open to date. In this paper, we discuss the pros and cons of each technique and present a direct quantitative comparison for a specific use case, using real data. Our results suggest that carefully combining the insights of problem modeling with state-of-the-art AI techniques provides significant advantages at reasonable complexity.  
\end{abstract}

\section{Introduction}
\input{introduction}
\section{Online Network Optimization (ONO)} \label{sec:ono}
\input{ono}

\section{ONO Techniques}
\input{techniques}

\section{Performance Evaluation}
\input{performance}

\section{Conclusions}
In this paper, we have stressed the need for automating online network optimization, and have examined several approaches such as online learning, data-driven optimization, and model-based AI techniques. We have performed a direct comparison of these techniques for the challenging problem of traffic steering and load balancing in dense, heterogeneous networks. Our study demonstrated how the inherent variability of network traffic can be successfully addressed by self-optimizing methodologies. The most prominent of them, is based on artificial intelligence, but also makes use of deep modeling insights of the problem. We have therefore demonstrated that the efficiency, robustness and scalability of online AI techniques can be substantially improved by applying the knowledge of the problem specifics.

\bibliographystyle{IEEEtran}
\bibliography{ONO_AI_references}

\section*{Biographies}
\textbf{Spyridon Vassilaras} is a principal researcher at the Huawei France Research Center since Dec. 2014. From 2003 till 2014 he was a researcher and professor at Athens Information Technology (AIT). He received the Dipl. Eng. degree from the National Technical University of Athens (1995) and the M.S. and Ph.D. degrees from Boston University (1997, 2001). His main research interests include machine learning and network optimization.  

\textbf{Luigi Vigneri} is a research scientist at IOTA Foundation. Previously, he was postdoctoral researcher at Huawei Technologies, France. He received his M.Sc. in Computer Engineering from Politecnico di Torino and Télécom ParisTech (2014), and his Ph.D. from EURECOM (2017). His main research interests include network optimization and distributed systems.

\textbf{Nikolaos Liakopoulos} received his B.S. in Physics and M.S. in Control and Computing from National Kapodistrian University of Athens, in 2012 and 2015. Since 2016 he has joined Huawei FRC, working towards his PhD in collaboration with Eurecom and UPMC. His research focuses in distributed and centralized control for wireless networks.

\textbf{Georgios S. Paschos} is a principal researcher at Huawei Technologies, France, since Nov. 2014. Previously, he held research positions at LIDS, MIT (USA),  CERTH-ITI and University of Thessaly, Greece, and VTT, Finland. He received his ECE diploma from Aristotle University of Thessaloniki 2002, and his PhD degree from University of Patras, 2006. 
He served as an associate editor for IEEE/ACM Trans. on Networking, IEEE Networking Letters, and as a TPC member of INFOCOM, WiOPT, and Netsoft. 

\textbf{Apostolos Destounis} is with Huawei since 2014, where he is now a Senior Research Engineer. During 2011-2014 he was with Alcatel-Lucent Bell Labs France. He received his  Ph.D. degree from Supelec in 2014, the M.Sc. from Imperial College London in 2010 and the Dipl. Eng. Degree from National Technical University of Athens in 2009. His research interests include optimization and machine learning applied to communication systems. 

\textbf{Thrasyvoulos Spyropoulos} is an Assoc. Professor at EURECOM, France, since Oct. 2010. He has also been a postdoctoral researcher at INRIA, Sophia-Antipolis (2006-2007) and Senior Researcher and Lecturer at ETH, Zurich (2007-2010). He received the Dipl. Eng. degree from the National Technical University of Athens (2000), and the Ph.D. degree from the University of Southern California (2006).

\textbf{M\'{e}rouane Debbah} entered the Ecole Normale Supérieure de Cachan (France) in 1996 where he received his M.Sc. and Ph.D. degrees. Since 2014, he is Vice-President of the Huawei France R\&D center and director of the Mathematical and Algorithmic Sciences Lab. He is an IEEE Fellow and a WWRF Fellow and has received more than 16 best paper awards. 
He has managed 8 EU projects, 24 national and international projects and received more than 16 best paper awards.

\end{document}

%% file: introduction.tex
\subsection{Modern Network Management and the Need for Online Optimization}

The high complexity of the existing management methodologies for wireless networks, together with the need for rapid network reconfiguration to maintain good performance, led in the past to the concept of Self-Organized Networks (SON), which proposed to automate network management \cite{Brunner09}. Although SONs are on the research spotlight for several years, their practical deployment is not yet fully realized. Moreover, a number of originally envisioned SON techniques have been outpaced by recent developments in 5G wireless and cloud-based networks, such as softwarization (e.g., Software Defined Networks - SDNs), Network Function Virtualization (NFV), and slicing. This evolution has promoted NFV Management and Orchestration (MANO) as a concept of re-designing network management altogether. With a growing attention on Artificial Intelligence (AI)-based methodologies to improve and automate MANO, the topic of SON has been revitalized \cite{AI_SON,Klaine17}. For example, major telecom operators announced recently their intention to employ AI methodologies for SON features, aiming to halve the times for routine network management tasks \cite{vodafone-trial}. Nevertheless, MANO and associated AI techniques do not cover the transport network or Radio Access Network (RAN). 

A fundamental goal of automated network management is to optimize the network. A number of key mobile network functionalities give rise to hard optimization problems in this context, including but not limited to: interference management, load balancing and traffic steering, energy saving, and more recently Service Level Agreement (SLA) enforcement for 5G network slices and services. While a number of works exist addressing such problems, the increasing \emph{spatiotemporal variability of network conditions} gives rise to complex \emph{online optimization} variants that must further satisfy a number of challenging requirements: (i) \emph{adaptability} to changing network conditions, (ii) \emph{agility}, to ensure adaptation occurs in a timely manner, and with low network overhead, and (iii) \emph{efficiency}, to ensure the network operating point remains as close to the theoretically optimal, i.e., an oracle-derived ideal configuration for the exact current network conditions and traffic demand.

\subsection{Variability of Network Traffic}

Telecommunication networks exhibit an inherent variability at many different levels and time scales. For example, optical fibers fail unexpectedly, the quality of wireless channels fluctuates rapidly due to multi-path fading, etc. While network management traditionally monitors and reacts to potential network failures, \emph{traffic demand is emerging as a key source of spatiotemporal variability} in modern wireless networks. This is partly due to increased network densification; each (small) cell deals with much fewer data users, not giving rise to the type of law of large number effects ``smoothing'' traffic demand in traditional macrocells. A second key reason is the growing use of smartphones and tablets as a user's main Internet-access device, making traffic demand follow content popularity ebbs and flows. 

These factors lead to frequent and often large fluctuations in traffic demand, over both time and space, affecting the majority of aforementioned mobile functions and related optimization problems. Due to the inherent traffic variability, maintaining good performance in the above features requires to \emph{continuously} re-solve the same complex optimization problem under different inputs (i.e., network and traffic conditions). Furthermore, since traffic surges might occur, the real performance of the system does not always follow the one predicted by the optimization model. Hence, Online Network Optimization (ONO) must satisfy two, often contradicting goals: (i) ensure close-to-optimal performance for the majority of predicted traffic conditions; (ii) ensure good (or at least stable) performance when the actual traffic conditions diverge from predicted ones. Let us clarify that ONO is restricted to time-scale granularity in the order of minutes or more, as network parameters reconfiguration incurs considerable overhead. Low time granularity mechanisms such as queue-based and packet-based agile control can be applied simultaneously in low network layers to cope with variability at finer time scales.  

\subsection{Data Traffic Prediction and Adaptation Methods}

There are many ways to capture the variability of traffic demand: in this study, we consider a generic model where $\lambda_{x,t}$ captures the number of service units demanded from location $x$ at time $t$ (depending on the problem setup, this could be bits/sec, packets/sec, flows/sec, etc.). In order to achieve the above strict goals of ONO in a modern mobile network, it is important to understand demand fluctuations and try to predict them. A number of traditional and more modern methods can be applied to this end, as we will detail subsequently.

Hence, one could apply a two-step approach to ONO: first, regularly predict the traffic demand in the next period(s) of interest, and then plug these estimates into the respective optimization model which will provide the optimal configuration for the predicted traffic demand. While such a method is well understood, its success hinges on correctly modeling an increasingly complex network and finding the appropriate prediction method that best ``fits'' each problem. Furthermore, robustness to prediction inaccuracies require keeping track of higher order statistics of demand variability and reformulating existing problems to hedge against prediction errors~\cite{Liakopoulos17}.

More recently, Artificial Intelligence (AI) methods, e.g., based on deep learning, have found significant success in solving complex pattern recognition and control problems. As a result, AI-based methods have attracted interest by the 5G community, offering a radically different way of approaching standard as well as new mobile network functionalities. Two are the key driving factors behind this trend: first, the abundance of large numbers of network data (known as ``Big Data'') that can be collected by SDN-enabled network components at various network layers; second, the fact that AI methods are usually ``model-free''. AI-based methods for ONO essentially bundle the above two steps, prediction and optimization, into one. 

Nevertheless, due to the infancy of these methods for mobile networking problems, a number of important questions remain unanswered: (i) Which method performs better, and for which types of problems? (ii) How should such methods be best applied to mobile networking problems? (iii) What is the computational complexity and related convergence properties? Since the mathematical tools and the studied regimes vary widely among these methodologies, a theoretical comparison is often out of reach. Therefore, the important question ``\emph{how should online network optimization be applied in 5G and beyond networks?}'' remains open to date. 

\subsection{Comparing ONO Techniques and the Case for Problem-Adapted AI}

To obtain some initial answers to these questions, we revisit the topic of AI for SON. Our main contributions can be summarized as follows:
\begin{itemize}
\item We propose the framework of ONO, which includes various machine learning and optimization techniques, aiming to predict traffic variability, and optimize networks by adapting to it in an online manner. 
\item We highlight the difficulty of comparing online optimization techniques, and provide intuition into what the strong features for each technique are.
\item We focus on the usecase of \emph{traffic steering} for dense, heterogeneous networks to perform a detailed quantitative comparison of various techniques. Our results suggest that traditional two-step approaches to ONO suffer from a trade-off between robustness and performance. On the other hand, we show that AI techniques rapidly adapt to changing network conditions, ensuring high accuracy.
\item Finally, we demonstrate that direct application of AI techniques to mobile networking problems suffers from state space explosion, requiring the tuning of a prohibitively large number of parameters. Instead, we show that it is possible to use domain knowledge and re-formulate the learning of a different set of parameters (smaller in size) which suffice to optimize the network performance. This makes a strong case for \emph{Problem-Adapted Artificial Intelligence} for future wireless network management. 
\end{itemize}

%% file: ono.tex
\subsection{Architecture \& Platform} 
\begin{figure*}[h]
  \begin{center}
		\includegraphics[width=0.67\textwidth]{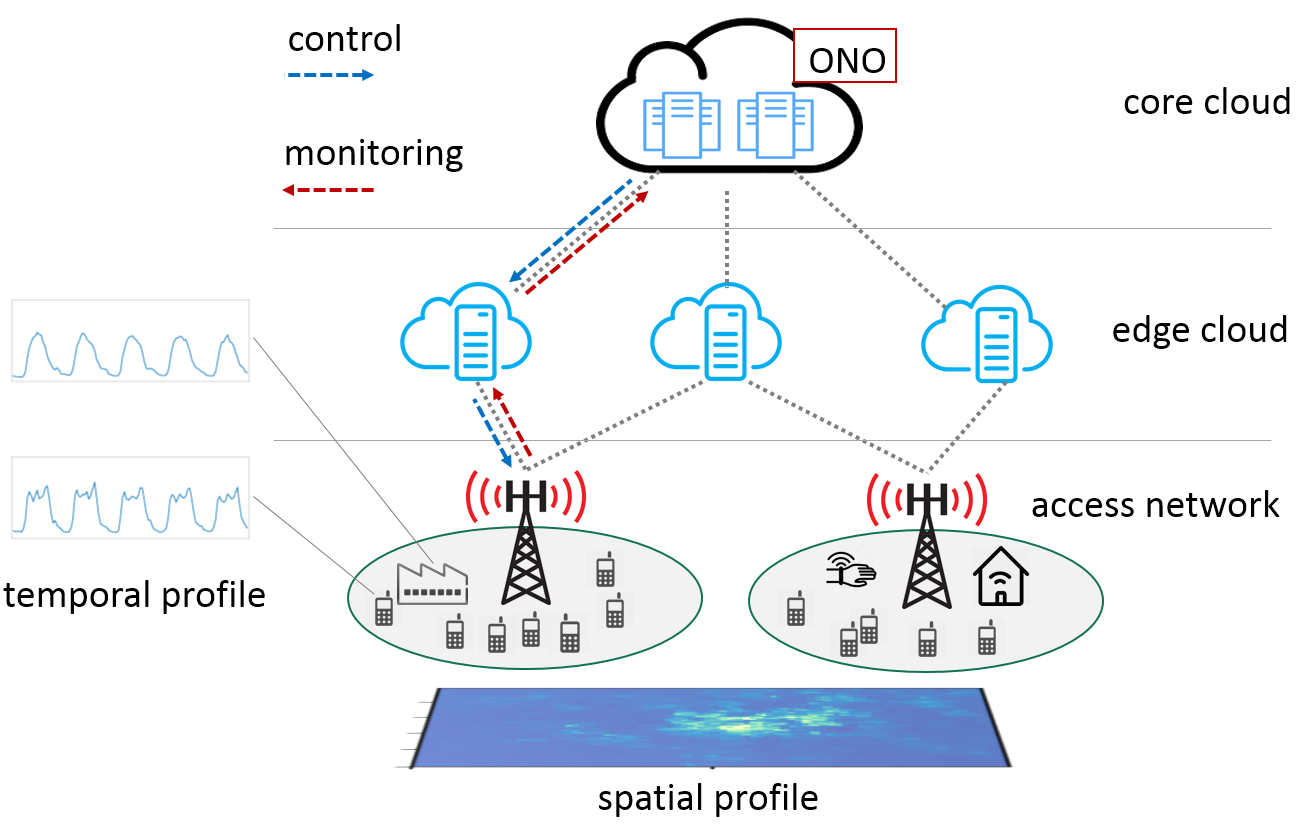}
    \caption{Architecture of a mobile network with ONO.}
		\vspace{-5.5mm}
    \label{fig:architecture}
  \end{center}
\end{figure*}

Online network optimization repeatedly adapts the network configuration in order to optimize performance, as measured by various key performance indicators. In this paper, we consider ONO in future wireless networks and adopt the high-level network architecture shown in Fig.~\ref{fig:architecture}. In such networks, traffic will be generated from / destined to a huge number of different mobile devices, including smartphones, Internet of Things and other industrial end-usage devices. We assume that applications as well as VNFs will run on generic hardware hosted in various data centers collectively referred to as the cloud. These data centers can be categorized as edge cloud (often small server installations distributed over a geographical area), and core cloud (large server installations typically serving a much larger area). A crucial architectural assumption for ONO is the real-time availability of monitoring data (facilitated by standardization efforts such as 3GPP's NWDA and ETSI's ENI SG \cite{Kibria18}) and the remote controllability of the access network (enabled by technologies such as SDN and NFV). In the architecture shown in Fig.~\ref{fig:architecture}, the ONO controller rests at a central location, though distributed architectures may be envisaged.

\subsection{Performance Metrics}

Network performance is a very broad term and needs first to be quantified in order to be optimized. Additionally, in ONO we are not just interested to optimize a single localized metric at a given instance but are rather concerned with control of \emph{network-wide} parameters over various time scales to continuously adapt to changing user demand and operating conditions. For instance, the achievable data rate in the downlink channel of a radio cell varies over time and from user to user.

\textbf{Statistical Performance Metrics:} While one might try to optimize the instantaneous data rate of every user at every time instance, network-wide optimization and management can only occur at a longer time scale, and thus statistical metrics are usually chosen. These include:
\begin{itemize}
\item{\emph{Time average:} of the metric of interest often also averaged over many connections, links, cells, etc. Popular variants include averaging over a time window or using a forgetting factor for old samples.} 
\item{\emph{Variability:} captured by variance or higher order statistics.}
\item{\emph{Compliance probability:} the probability that the metric exceeds / falls short of a given threshold.}
\item{\emph{Regret:} a measure of accumulated ``error'' of the algorithm, compared to an ideal or oracle one (often not implementable in practice).}
\item{\emph{Fairness:} utility that takes into account the relative value for the metric of interest across different  entities (e.g., users), often generalized by ``alpha-fairness''; the latter parameterizes the trade-off between fairness and optimizing the average metric over all users.}
\end{itemize}

However, more often than not, ONO is concerned with optimizing a number of heterogeneous and even contradicting metrics such as network operating cost and user Quality of Service (QoS), connection throughput and delay, etc. There are three common approaches to tackle such cases: 
\begin{itemize}
\item{Combining the different metrics into one (e.g., as an appropriately weighted sum).}
\item{Applying multi-objective optimization and provide Pareto optimal solutions.}
\item{Optimize for one metric (or a weighted sum of a subset of the metrics) and use the remaining metrics as constrains, e.g., optimize throughput subject to a maximum allowable delay constraint.} 
\end{itemize}
 
\textbf{Algorithmic Metrics:} One of the difficulties in comparing ONO algorithms developed independently is that they are often concerned with optimizing a different metric. In addition, when comparing different ONO algorithms, there are numerous criteria to consider, such as algorithmic complexity/scalability, training time, required training data size, sensitivity to noisy training data, response time, convergence time, convergence reliability, control cost, and communication overhead. A detailed discussion of some of the above criteria, from the point of view of machine learning algorithms, is provided in \cite{Klaine17}. 

\subsection{ONO Usecase: Traffic Steering}

Mathematically, a typical ONO problem consists in repeatedly optimizing a function $f(\boldsymbol{\lambda}_t,\boldsymbol{\pi}_t)$ where $\boldsymbol{\pi}_t$ is a vector of control variables and $\boldsymbol{\lambda}_t$ is a vector random process. If $\boldsymbol{\lambda}_t$ was perfectly known, finding the optimizing $\boldsymbol{\pi}_t$ would be identical to solving a standard optimization problem. However, $\boldsymbol{\lambda}_t$ is unknown and only prior values of it ($\boldsymbol{\lambda}_0, \boldsymbol{\lambda}_1, \ldots , \boldsymbol{\lambda}_{t-1}$ - in the discrete time case) have been observed. Knowledge of the statistical properties of the random process might be available as well. 

Note that we have used the same symbol to express the unknown variables $\boldsymbol{\lambda}_t$ in function $f(\boldsymbol{\lambda}_t,\boldsymbol{\pi}_t)$ and the varying traffic demand measure $\lambda_{x,t}$ in the previous section, as the prime source of network variability is the fluctuation of traffic load. If we observe a specific location $x_0$, $\{\lambda_{x_0,t}\}$ for all $t$ becomes a scalar time series with seasonal structure. Figure~\ref{fig:architecture} shows the weekly traffic fluctuation in two different places in Milan as measured by Telecom Italia in 2014 \cite{telecom_italia}. We notice the 24-hour periodic ``diurnal pattern'', a common characteristic of all networks attributed to the daily human activity. 

On the other hand, when fixing a time instance $t_0$, $\boldsymbol{\lambda}_{t_0}$ (at all $x$) provides a spatial depiction of traffic demand in the network, as shown on the bottom of Figure~\ref{fig:architecture}. However, when considering the spatio-temporal properties of $\{\boldsymbol{\lambda}_{t}\} \!$~, a number of new characteristics appear. When the city population is moving from the outskirts to the city center during the morning commute, strong negative traffic correlations couple the traffic evolution of different locations. In contrast, night and day traffic demand are positively correlated over space (i.e., most locations have low traffic in the night and high traffic in the day). All such correlations, if understood, can improve prediction accuracy.

For reasons of exposition, we turn our focus on the following example application of the ONO framework. From the point of view of a mobile network operator, we would like to use ONO techniques to decide how to steer user traffic to different Base Stations (BS). Due to the spatiotemporal fluctuations of user demand, the incurred load at each BS also fluctuates, affecting the QoS received by the users. The network operator should optimize the steering of mobile traffic to available BSs in order to keep the loads balanced, and improve overall user experience.

Our scenario is based on the Telecom Italia dataset \cite{telecom_italia}, and has the following ingredients:
\begin{itemize}
\item The dataset determines $\lambda_{x,t}$ in 10,000 grid locations in Milan, in intervals of ten minutes, for the duration of five weeks.
\item The control variables $\pi_{x,t,j}\in [0,1]$ indicate what fraction of traffic $\lambda_{x,t}$ is steered to BS $j\in\{1,2,\ldots,K\}$, where $K$ is the number of BSs in the area. 
\item The load contribution to BS $j$ from location $x$ is given by $\pi_{x,t,j}\lambda_{x,t}/R_{x,j}$, where $R_{x,j}$ is the connection rate from BS $j$ to location $x$. The total BS load $\rho_j(\boldsymbol{\lambda}_t, \boldsymbol{\pi}_t)$ of BS $j$ is found by summing up the contribution from each location.
\item At each time instance, the goal of our ONO is to minimize the average cost over all BSs, where the cost at BS $j$ is given by $-\log (1-\rho_j)$. This objective leads to proportionally fair loads.
\end{itemize}

%% file: techniques.tex
Next we present 3 different techniques for solving the typical ONO problem of optimizing $f(\boldsymbol{\lambda}_t,\boldsymbol{\pi}_t)$ when $\boldsymbol{\lambda}_t$ is uncertain.  

\subsection{Data-Driven Optimization based on Forecasts}

Under the assumption that if $\boldsymbol{\lambda}_t$ was perfectly known solving for the optimal $\boldsymbol{\pi}_t$ would be straightforward, the two-step approach of first predicting $\boldsymbol{\lambda}_t$ and then determining $\boldsymbol{\pi}_t$ seems the most natural one. 
In the case $\boldsymbol{\lambda}_t$ represents traffic, the diurnal patterns of traffic demand can be accurately predicted by advanced time series analysis techniques, such as the seasonal AutoRegressive Integrated Moving Average (ARIMA) model, which remove the periodicity and trend from the signal. More recently, with the proliferation of neural networks, alternative techniques have been developed with increased efficiency, most notably \cite{HTM}: Recurrent Neural Nets (RNN), Hierarchical Temporal Memory (HTM), Long Short-Term Memory (LSTM). LSTM is particularly good at multivariate time-series analysis \cite{LSTM}, and therefore very useful when traffic exhibits spatial correlations on top of temporal correlations, as in our use case. When the associated optimization problem with known $\boldsymbol{\lambda}_t$ is hard, approximation algorithms and program relaxations can be used to provide a ``good'' solution in a reasonably small time.  

The quality of the prediction is crucial for the actual performance of this approach. If predictions by an oracle were available, the solution to the associated optimization problem would give the best configuration at any time instance. In reality, prediction is never perfect and thus we need to take prediction error into consideration. After a good forecasting tool is applied to our data, we may assume that the prediction error can be described by an i.i.d. Gaussian process $N_{x,t}$, i.e., ${\lambda}_{x,t}=\hat{\lambda}_{x,t}+N_{x,t}$ , where $\hat{\lambda}_{x,t}$ is the predicted value. This motivates us to optimize for the actual traffic ${\lambda}_{x,t}$, which in our model is an i.i.d. stochastic process.  Such stochastic optimizations can be efficiently solved with Robust Data-Driven techniques, such as those in \cite{Bertsimas14}. Later in the paper, we will provide numerical evidence on how the accuracy of prediction affects the performance of ONO techniques, as higher prediction accuracy reduces the variance in $N_{x,t}$.

A method for performing robust, prediction-driven ONO for traffic steering is described in \cite{Liakopoulos17}. 
There, the resulting stochastic optimization problem is configurable to protect an SLA on the average queue delay experienced in the BS queues as well as guaranteeing service with a probability $\epsilon$ (a design parameter), while it optimizes the expected load and prediction error variance. Most importantly, the problem is shown to be convex and efficiently solvable with the proposed algorithm.

\begin{figure*}[t]
	\begin{center}
	\includegraphics[width=0.75\textwidth]{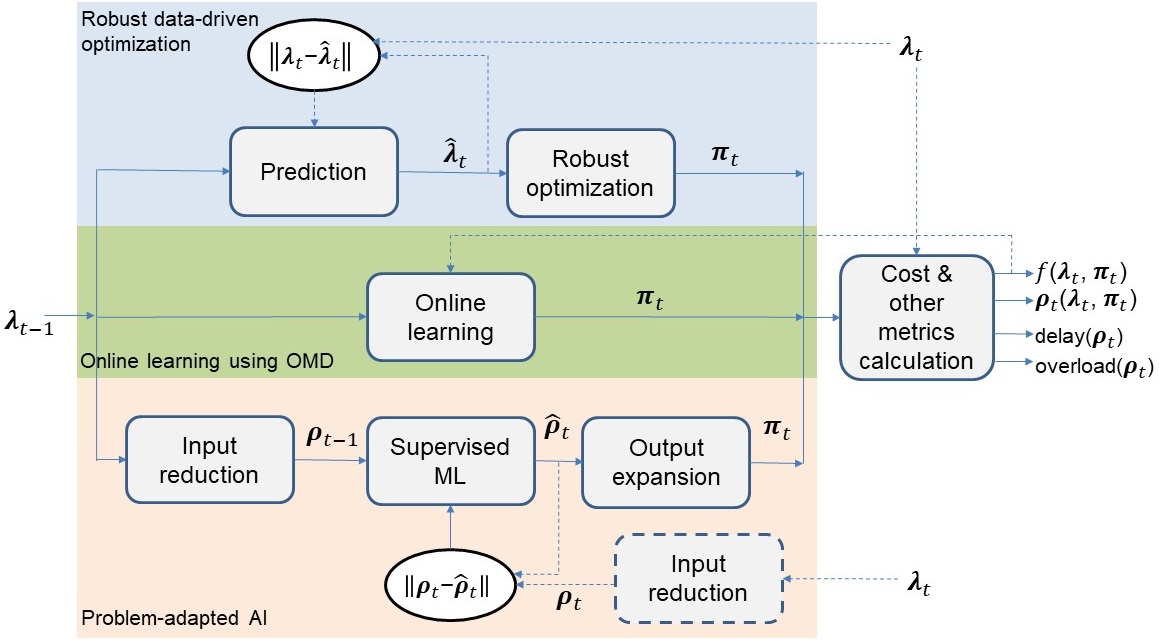}
	\caption{Diagram of the simulator modules and their interconnections. Solid lines depict the flow of data during time slot $t$ while dashed lines show the feedback loop after the actual process realization at time $t$ is revealed. Feedback loops can be used either offline or online for training/updating the ML components and performing OMD in online learning. The two instances of the \emph{Input reduction} module are performing the same operation at different times. This operation is problem specific, and in our case consists of first calculating the optimum $\boldsymbol{\pi}^*(\boldsymbol{\lambda}_{t-1})$ for the observed $\boldsymbol{\lambda}_{t-1}$ and then computing $\boldsymbol{\rho}(\boldsymbol{\lambda}_{t-1},\boldsymbol{\pi}^*_{t-1})$ (and similarly for the feedback loop with input $\boldsymbol{\lambda}_{t}$).}
	\label{fig:simulator_flow_chart}
	\end{center}
\end{figure*}

\subsection{Online Learning}\label{Online_learning}

When data monitoring is complex, or training over the entire dataset is computationally infeasible, it is preferable to use techniques that directly adapt the decisions on the most recently observed data. This draws from \emph{online learning}, a machine learning method used to answer a \textit{sequence} of questions given data available in a sequential order~\cite{hazan2016}. Unlike the two-step approach, the online learning algorithm provides a suggested control $\boldsymbol{\pi}_t$ at round $t$ of the sequence, chosen from a set of feasible controls without explicitly predicting $\boldsymbol{\lambda}_t$, but rather from available previous values of the control and uncertain parameters. 
Let us define the \emph{loss} at time $t$, as the discrepancy between the value of $f$ under the suggested control and the optimal control under perfect knowledge of $\boldsymbol{\lambda}_t$. The learner's goal is to minimize the cumulative loss by adapting its suggested controls based on already observed losses. 

A special case of interest is when the control set and the loss function are both convex \cite{hazan2016}. A prominent
algorithm in this case is \emph{online gradient descent}, where at each iteration of the sequence we take a step from the most recent control in the direction of the negative of the gradient of the previous loss~\cite{zinkevich2003}. \emph{Online Mirror Descent} (OMD) has been proposed as a powerful generalization of the online gradient descent method. At a high level, in OMD a point in the primal space is mapped to the dual space through a function (called mirror map) to perform the gradient update. It has been shown that by picking a proper mirror map, the regret of OMD grows as $\sqrt{\log(d)}$ (where d is the dimensionality of the problem) and hence it is nearly-independent of $d$~\cite{hazan2016}. We apply OMD in our illustrative traffic steering example, based on the entropic mirror map which fits well to the geometry of the corresponding optimization problem.

\subsection{Problem Adapted Artificial Intelligence}

In the first ONO technique described above, AI methods can be used in the prediction step of the two-step ONO approach. A single-step approach can be taken instead, where a neural network can be trained to infer the optimal decision $\boldsymbol{\pi}_t$ given the past traffic demands $\boldsymbol{\lambda}_1, \boldsymbol{\lambda}_2, \ldots, \boldsymbol{\lambda}_{t-1}$. 
Overall, the potential of AI in traffic control has been underexplored. One approach is to use Reinforcement Learning: start with no knowledge of the system and progressively train a deep neural network by observing the outcomes of the policies applied \cite{Alizadeh}. Recent work \cite{TE_withDRL} has applied this idea for traffic routing.

Here, instead, we propose to leverage the cyclostationary behavior of traffic and the fact that optimal solutions to past traffic instances can be computed offline to train the neural network in a \emph{supervised} manner: first, in order to reduce the dimensionality, we set the output to the optimal BS loads (40 in our example), and use the predicted optimal loads to recover the optimal traffic steering (10000 locations, each with 40 control variables, for a total of 40000 control variables) through analytical models such as in \cite{deVec}. Then, each of the past samples is ``labeled'' with the optimal BS loads and used as the training set. 
Once training is completed, the trained neural network takes as input $\boldsymbol{\rho}_{t-1}$ and makes a prediction $\hat{\boldsymbol{\rho}}_{t}$ of the optimal loads. Based on $\hat{\boldsymbol{\rho}}_{t}$, through the models of \cite{deVec}, the corresponding steering $\boldsymbol{\pi}_{t}$ is derived. 
In our traffic steering case study, we have used a neural network based on LSTM. Such networks are equipped by a state which acts as memory and use gates to regulate the importance of previous versus current data samples \cite{LSTM}. This architecture allows them to capture temporal short and long-range dependencies in the data, making them ideal for our scenario. 

We emphasize that the reduction of the state space contributes significantly to the practicality of the whole approach. Note that the proposed state space reduction is system and problem specific and does not rely on feature selection. A key message of this paper is that in most ONO settings such \emph{problem-adapted AI} techniques can be proven greatly superior to blindly applying a deep neural network, often making a prohibitively complex problem tractable.

%% file: performance.tex
In this section, we report the performance results of different ONO approaches to the problem of traffic steering, where the goal is to minimize the average cost defined in section \ref{sec:ono}
such that BSs are not overloaded (i.e., $\rho_j<1$, for all $j$). 
To perform such analysis, we have built a time-slotted simulator in Python, driven by real data sets, with the long-term vision to provide a tool that will allow to easily compare different ONO techniques. 

\subsection{The simulator}
We have developed our simulator in a modular manner so that it can be extended and modified to include other ONO techniques (or variants of the already implemented ones) and simulate different use-cases. It currently covers the 3 techniques described in the previous section.     

The simulator is composed by the following main modules (Fig.~\ref{fig:simulator_flow_chart}):
\begin{itemize}
\item \textit{Prediction module}: As part of the robust data-driven optimization technique, the prediction module is performing a prediction of the next value of $\boldsymbol{\lambda}_t$ given its previous value(s). Five different versions of this module have been implemented and can be used in the simulator without modifying any other part of it, thanks to its modular design: sample mean, previous value, seasonal ARIMA, LSTM, and oracle. The last one, relies on a priori knowledge of future traffic demands, which is not realistic in practice but is used to compare the performance of our algorithms to the performance that could be achieved by perfect prediction. 
\item \textit{Robust optimization module}: This module  takes as input the output of the prediction module and performs robust ONO as described in \cite{Liakopoulos17}.
\item \textit{Online learning module.} This module is the main module of the Online Learning technique, currently implementing OMD based on the entropic mirror map. Different mirror maps or simple gradient descend algorithms could be implemented in this module.
\item \textit{Input reduction module}: This module combines with the AI module to implement the problem-adapted AI technique. For the traffic steering problem, this module translates between $\boldsymbol{\lambda}$ and $\boldsymbol{\rho}$ as explained in the previous section.

\item \textit{Output expansion module}: This module combines with the AI module to implement the problem-adapted AI technique. For the traffic steering problem, it translates between $\boldsymbol{\rho}$ and $\boldsymbol{\pi}$ as explained in the previous section. 
\item \textit{Supervised machine learning module}: This module applies some AI technique to learn the best response in the reduced output space given an input in the reduced input space. The currently implemented supervised ML model is a LSTM with 2 hidden layers.
\item \textit{Cost and other performance metrics calculation module}: This module calculates the instantaneous and time averaged costs as well as other key performance metrics, such as average delay and percentage of rejected traffic due to BS overload.
\end{itemize}

\subsection{Performance evaluation and comparison}

\begin{figure*}[h]
	\centering
	\includegraphics[width=0.65\textwidth]{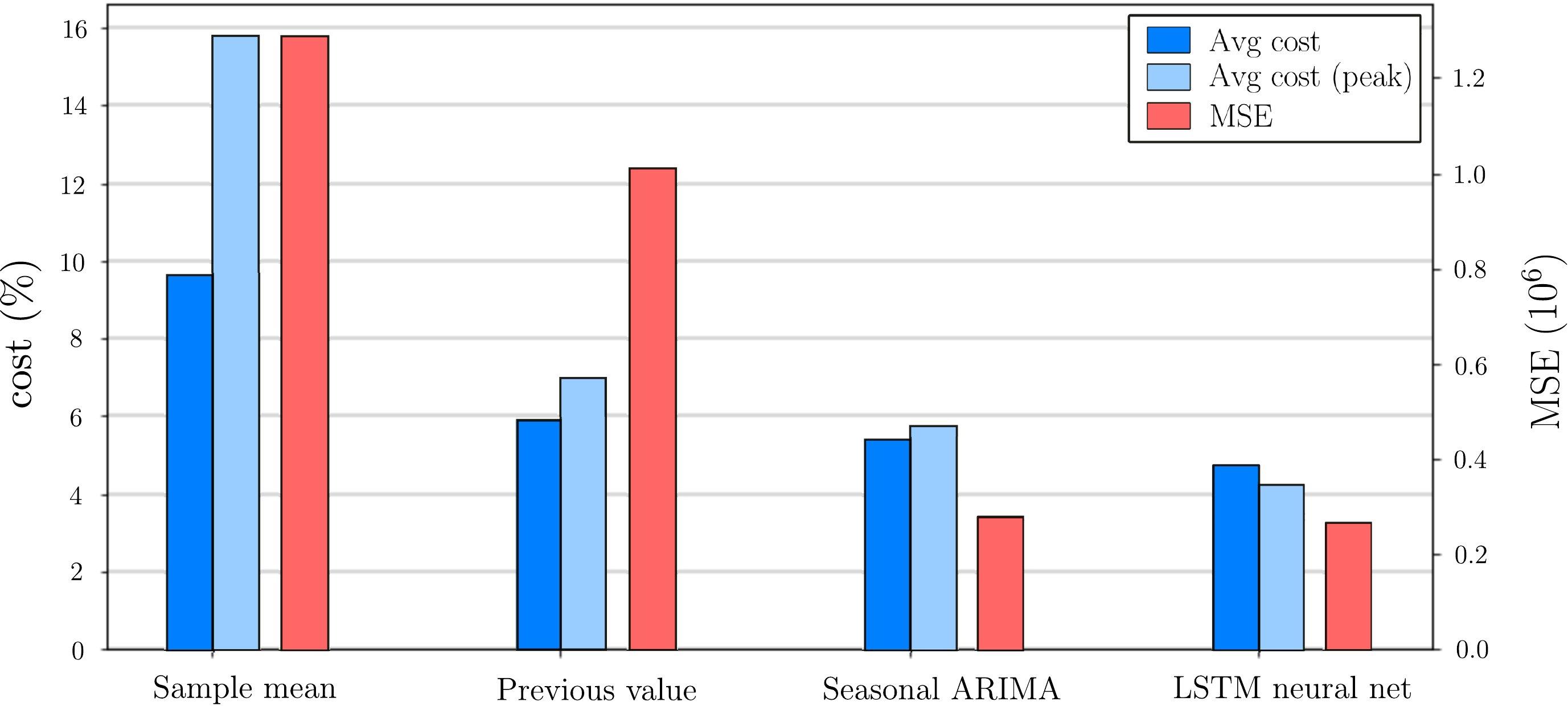}
	\caption{Comparison between traffic prediction methods as inputs for data-driven optimization. Cost is averaged over all BS and timeslots ignoring timeslots when cost becomes infinite. We depict the average cost and the average cost during peak hours, as a percentage of the distance from the average cost achieved by perfect prediction. Mean squared error (MSE) values between the predictions of each predictor $\hat{\boldsymbol{\lambda}}_t$ and the actual value $\boldsymbol{\lambda}_t$ are also shown. The ``sample mean'' predictor uses the empirical mean of the past observed values of traffic at the same day and hour in previous weeks. It also provides the associated variance for the purposes of robust optimization.}
	\label{fig:prediction}
\end{figure*}

The performance evaluation results are obtained by feeding our simulator with the Milan dataset and associated network topology. We start by a comparison of the different versions of the prediction module used in the robust data-driven optimization technique. Fig.~\ref{fig:prediction} shows how much worst is the average cost achieved by robust optimization using each one of the 4 realistic versions than the cost ideally achieved by perfect prediction (oracle). The same Figure also depicts the MSE between the predictions of each predictor and the actual value. We observe that LSTM achieves the best performance (only about 4\% worst than the theoretical optimum) closely followed by seasonal ARIMA. As expected, better prediction accuracy leads to lower average cost and the improvement is more pronounced for peak hour cost. 

\begin{figure*}[h]
	\centering
	\includegraphics[width=0.59\textwidth]{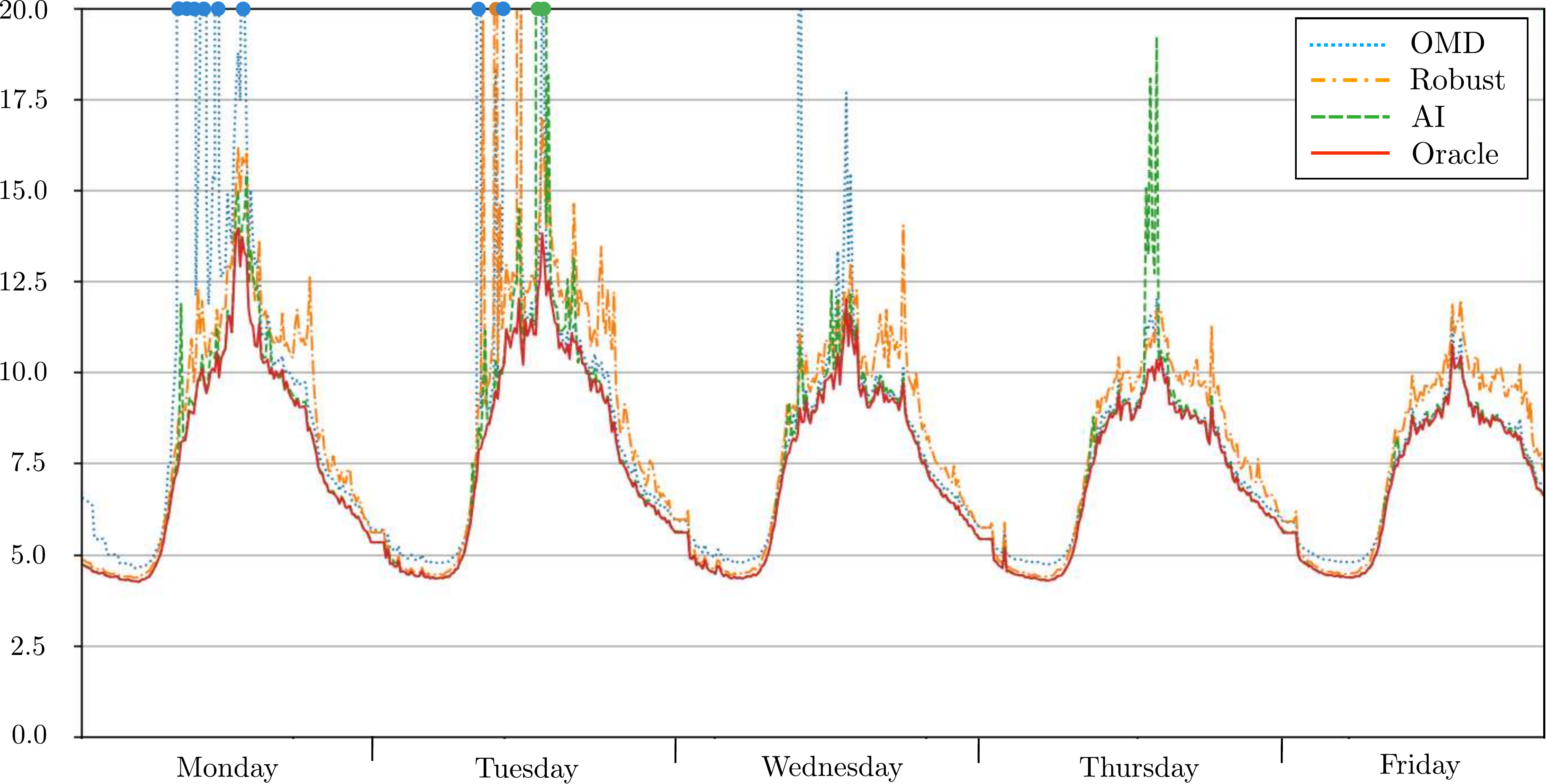}
	\caption{Average delay (in \textit{ms}) over all BSs during 5 days with different ONO techniques. Traffic prediction for robust optimization is performed using the LSTM predictor. Filled circles represent infinite delay, in which case part of the incoming traffic demand is rejected.}
	\label{fig:average_delay}
\end{figure*}

\begin{table*}[h]
	\renewcommand{\arraystretch}{1.3}
	\centering
	\caption{Summarized statistics per day and for the whole period: (i) average delay in \textit{ms} over all BSs at daytime (i.e., between 7:30 and 20:30); (ii) mean squared error (MSE) of average delay over all BSs against the optimal delay; (iii) rejected traffic in percentage of the total traffic. To compute average delay and MSE values, time slots with infinite delay are omitted. We highlight the lowest value (per day and overall) in bold.}
	\label{tab:sim_outcome}
	\begin{tabular}{|l|l|ccccc|c|} \hline
		\rule{0pt}{2.5ex}  	& \textbf{Method}	& \textbf{Mon}	& \textbf{Tue} 	& \textbf{Wed} 	& \textbf{Thu} 	& \textbf{Fri} 	& \textbf{Avg} 	\\ \hline \hline
		\parbox[t]{2mm}{\multirow{4}{*}{\rotatebox[origin=c]{90}{\textbf{Avg delay}}}}
		& Robust data-driven optimization			& 10.95			& 11.90			& 10.01			& 9.50			& 9.41			& 10.35			\\ 
		& Online learning using OMD				& 11.62			& 11.50			& 9.73			& \textbf{8.79}	& 8.52			& 10.03			\\ 
		& Problem-adapted AI 				& \textbf{9.90} & \textbf{10.09}& \textbf{9.13} & 9.07			& \textbf{8.43}	& \textbf{9.32} \\
		& \textit{Oracle}	& \textit{9.57}	& \textit{9.81}	& \textit{8.86}	& \textit{8.54}	& \textit{8.35} & \textit{9.02} \\ \hline \hline
		\parbox[t]{2mm}{\multirow{3}{*}{\rotatebox[origin=c]{90}{\textbf{MSE}}}}
		& Robust data-driven optimization				& 2.30			& 17.23			& 1.61			& 0.95			& 1.21			& 4.66			\\
		& Online learning using OMD			& 8.63			& 21.66			& 9.29			& \textbf{0.18} & 0.04			& 7.96 			\\ 
		& Problem-adapted AI				& \textbf{0.49}	& \textbf{0.95}	& \textbf{0.25}	& 2.63			& \textbf{0.02}	& \textbf{0.87}	\\ \hline \hline
		\parbox[t]{2mm}{\multirow{3}{*}{\rotatebox[origin=c]{90}{\textbf{Rejected}}}}
		& Robust data-driven optimization			& 0.00			& 0.06			& 0.00			& 0.00			& 0.00			& \textbf{0.01}	\\ 
		& Online learning using OMD				& 1.10			& 0.24			& 0.00			& 0.00			& 0.00			& 0.27			\\ 
		& Problem-adapted AI				& 0.00			& 0.26			& 0.00			& 0.00			& 0.00			& 0.05			\\ \hline
		
		 \hline
	\end{tabular}
\end{table*}

In order to compare the performance of the 3 ONO techniques, we depict the resulting delay (averaged over all BSs) over the course of 5 days in Fig.~\ref{fig:average_delay}. Note that in the event of a BS overload, the delay becomes infinity, and the incoming traffic associated to this BS is rejected. According to our dataset, the traffic is heavier on Monday and Tuesday compared to the rest of the week (we suspect this is due to the imminent Sant'Ambrogio holiday in Milan) making this specific week an ideal setting for comparison of ONO techniques. Table~\ref{tab:sim_outcome} presents average KPIs of applying the 3 ONO techniques on the same data. The experimental results shown on Fig.~\ref{fig:average_delay} and Table~\ref{tab:sim_outcome} lead us to the following observations:

\textit{1) Robust optimization} can be tuned to trade-off average cost / delay against excessive overload and thus protect against SLA violations. In the simulated scenario, we configured the method to limit the percentage of rejected traffic to less than $0.1\%$, and this requirement is indeed satisfied (see Table~\ref{tab:sim_outcome}). However, the average delay is the worst among all three methods.

\textit{2) Online learning} adapts relatively quickly to a good state, though sometimes fails to react to a rapid increase in traffic. As OMD makes no assumptions on the statistics of traffic fluctuations, it is best suited for applications where $\boldsymbol{\lambda}_t$ is highly non-stationary. When $\boldsymbol{\lambda}_t$ follows a stationary process with seasonality (as in our case) alternative methods which make use of acquired knowledge on the statistics of the underline process are advantageous. However, as OMD does not apply any ``margin'' to guard against sudden traffic fluctuations, a steep traffic increase might lead to large volumes of rejected traffic. This is confirmed by our experiments on Monday and Tuesday when traffic radically changes from nighttime to morning hours. In contrast, steep traffic increases can be handled more easily when the average cumulative traffic is lower: during the last three days, OMD achieves delays close to the oracle with low or no constraint violations.

\textit{3) Problem-adapted AI} exhibits the lowest average delay among the 3 techniques while keeping rejected traffic at very low levels. It achieves this by directly predicting the optimal BS loads instead of predicting the traffic and calculating the ``optimal'' loads while trying to guard against traffic prediction errors. This way it both averages out a large number of small prediction errors and learns to be only as conservative as needed in its decisions. At the same time, unlike robust optimization, problem-adapted AI can scale with the number of locations, as explained in the previous section.     


\subsection{Which method to choose?}

Although the simulations results largely lean in favor of problem-adapted AI, the answer is not trivial. Surely, problem-specific AI provides the \textit{best tradeoff between robustness and performance} and scales better than robust optimization. However, its performance largely depends on the quality of training. Proper training requires good training data and a potentially huge amount of resources (i.e., memory, CPU) depending on the dimensionality of the problem, which are not always available. This is the reason why it is important to combine AI with modeling intuition in order to significantly reduce the problem state space. On the other hand, online learning algorithms are very simple and quick, but sometimes fail to protect the system from rapid traffic surges. Deciding which method to choose is therefore largely dependent on the application, the problem size and the available resources. 
